# Compact Multi-scale Periocular Recognition Using SAFE Features

Fernando Alonso-Fernandez, Anna Mikaelyan, Josef Bigun
Halmstad University. Box 823. SE 301-18 Halmstad, Sweden
{feralo, annmik, josef.bigun}@hh.se, http://islab.hh.se

*Abstract*—In this paper, we present a new approach for periocular recognition based on the Symmetry Assessment by Feature Expansion (SAFE) descriptor, which encodes the presence of various symmetric curve families around image key points. We use the sclera center as single key point for feature extraction, highlighting the object-like identity properties that concentrates to this unique point of the eye. As it is demonstrated, such discriminative properties can be encoded with a reduced set of symmetric curves. Experiments are done with a database of periocular images captured with a digital camera. We test our system against reference periocular features, achieving top performance with a considerably smaller feature vector (given by the use of a single key point). All the systems tested also show a nearly steady correlation between acquisition distance and performance, and they are also able to cope well when enrolment and test images are not captured at the same distance. Fusion experiments among the available systems are also provided.

## I. INTRODUCTION

Periocular biometrics has gained popularity in recent years as a stand-alone modality due to its suitability for relaxed and/or non-ideal conditions where traditional modalities like iris or face may struggle [2]. Among its advantages we can mention the availability over a wide range of acquisition distances where iris or face may not be available, the non-necessity of accurate iris segmentation, and a superior performance under image degradations.

This paper presents a new approach for periocular recognition based on the Symmetry Assessment by Feature Expansion (SAFE) descriptor proposed in [11] for forensic fingerprints. It has been also used for periocular recognition [10], but with a different setup. The SAFE descriptor encodes the presence of various symmetric curve families (Figure 1, top) in annular rings around key points. A novelty is that we employ as unique key point the sclera center, and the annular rings are defined starting from the sclera boundary. An advantage of using a single key point is that the feature vector is of very low size in comparison with other popular periocular descriptors. The proposed method is also able to cope efficiently with image rotations, since rotation compensation can be achieved in the feature space via multiplication with complex scalars. Our setup also highlights the unique object-like identity properties concentrated to the eye center of each individual, which can be discriminated with a reduced set of symmetric curves.

The system is tested against traditional baseline Local Binary Patterns (LBP) [12], Histogram of Oriented Gradients (HOG) [7] and Scale-Invariant Feature Transform (SIFT) [9]. We employ a database of periocular images captured with a digital camera at several distances. The baseline features are also improved with a choice of different metrics for the computation of matching scores. Our system achieves top performance w.r.t. baseline features, using a feature vector that is (in the less favorable case) 40% smaller than the other descriptors. Further performance improvements can be obtained by fusion of the available systems. We also analyze the impact of acquisition distance (i.e. image scale changes) in the recognition performance. Three of the matchers show a steady performance improvement as the acquisition distance is reduced, but one of them (SIFT) does not show a clear correlation. In addition, all the matchers are able to cope relatively well with differences in the acquisition distance between enrolment and test images.

## II. SYSTEM DESCRIPTION

An overview of the employed system is shown in Figure 2, with a detailed description given next.

### A. Orientation Field Via Symmetry Filters

The system starts by extracting the complex orientation map $h = (D_x + iD_y)^2$ of the image $f$ via symmetry derivatives of a Gaussian, $\Gamma^{\{n,\sigma^2\}}$ (with $D_x$, $D_y$ being the estimated partial derivatives of $f$):

$$\Gamma^{\{n,\sigma^2\}} = \begin{cases} (D_x + iD_y)^n g_\sigma(x,y) & (n \geq 0) \\ (D_x - iD_y)^{|n|} g_\sigma(x,y) & (n < 0) \end{cases} \quad (1)$$

The complex-valued orientation field is given by [6] $h = \left\langle \Gamma^{\{1,\sigma^2\}}, f \right\rangle^2$, with parameter $\sigma$ defining the size of the derivation filters used in the computation of image $h$, i.e. the inner-scale of interest. We employ different $\sigma$ in our experiments to capture features at different scales. Prior to the orientation field computation, we apply contrast enhancement via Contrast-Limited Adaptive Histogram Equalization (CLAHE) [18] to compensate variability in local illumination. We employ CLAHE since it is usually the preprocessing choice with images of the iris region [15].

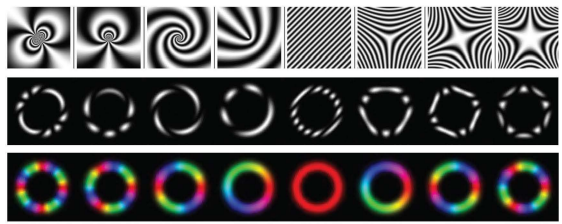

Fig. 1. *Top*: Sample patterns of the family of harmonic functions used as basis of the SAFE matcher (with *n*=-4:3). *Middle*: One pattern per original (top) but in selected ring support $|\psi_{nk}|$. *Bottom*: Filters $\psi_{nk}$ used to detect patterns above.

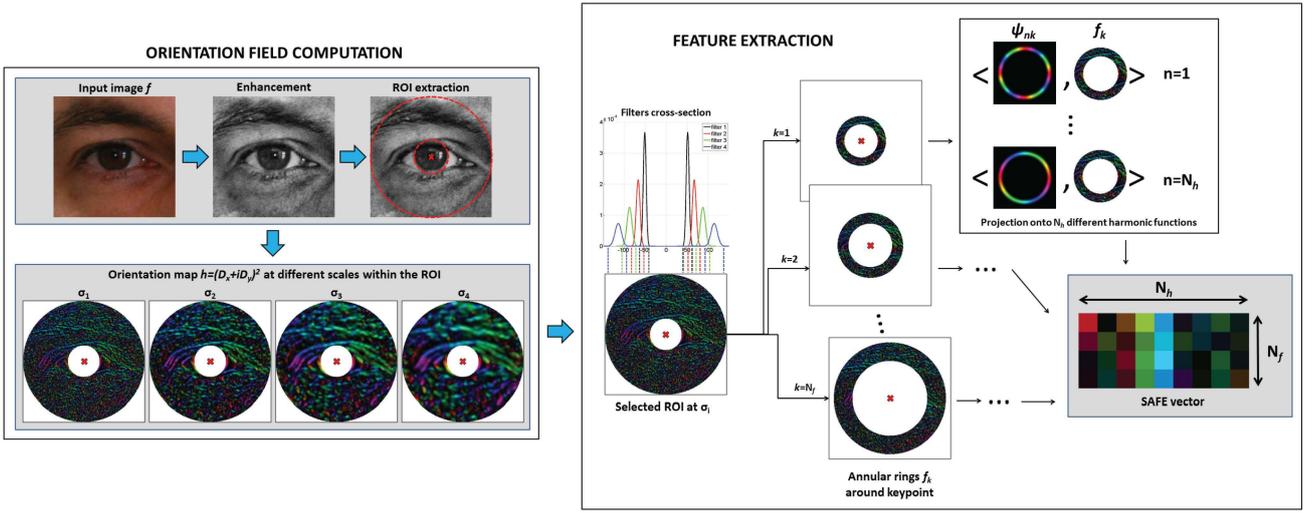

Fig. 2. SAFE periocular matcher. Left: complex orientation map and ROI extraction. Right: feature extraction process. Hue in color images of the orientation field encodes the direction, and saturation represent the complex magnitude.

### B. Orientation Field Descriptor

The feature extraction method describes orientation neighborhoods around key points by projection onto harmonic functions which estimates the presence of various symmetric curve families (Figure 1, top) around such key points. This is done in radially disjoint regions defined by concentric annular rings (Figure 1, middle), which is achieved with the modulus of the complex filter $\psi_{nk} = (1/\kappa_k) r^\mu e^{-\frac{r}{2\sigma^2}} e^{in\varphi}$ [10].

Here, $n$ defines the symmetry order (the pattern family to be detected, as shown in each column of Figure 1). Parameter $\mu$ defines the width of the filter, whereas the position of the filter peak is controlled by $\sigma$ via $r_k = \sqrt{\mu}\sigma$ (different than the $\sigma$ of Section II-A), with $r_k$ being the desired position. We can therefore tune the filters to a desired annular band, with known radius and width. For feature extraction, we define a range of radii $[r_{min}, r_{max}]$ around a key point, and build $N_f$ filters with peaks log-equidistantly sampled in this range (see right subplot of Figure 2). The normalization constant $\kappa_k$ ensures that $\|\psi_{nk}\| = 1$, so the filters have an underlying area of the same size and the value of extracted features is independent of the annuli size. In this work, we use the sclera center as unique key point. The annular band of the first filter (with peak $r_{min}$) starts at the sclera circle, and the band of the last filter (with peak $r_{max}$) ends at the boundary of the image. With this setup, the iris region is not processed, which from an anatomical perspective, does not belong to the periocular area [14]. In addition, using the sclera boundary as reference alleviates the effect of dilation that affects the pupil, because the sclera is not affected by such dilation. Moreover, the sclera is easier to detect than the pupil in visible range, which is the illumination that best suits the periocular modality, both in terms of performance and in terms of the impossibility of capturing near-infrared images in uncontrolled environments [2]. Additionally, periocular features become thereby fully complementary to iris features in multimodal applications.

Therefore, we use the result of scalar products of harmonic filters $\psi_{nk}$ with the orientation image $h$ to quantify the presence of pattern families as those shown in Figure 1 around a key point. We project $N_f$ ring-shaped areas of different radii around a key point onto a space of $N_h$ harmonic functions. The feature vector dimension describing a key point is thus given by an array **SAFE** of $N_h \times N_f$ elements. The elements $SAFE_{nk}$ are complex-valued and their magnitudes represent the amount of reliable orientation field within the annular ring $k = 1...N_f$ explained by the $n = 1...N_h$ symmetry basis.

### C. Matching

To match two feature vectors $\mathbf{SAFE}^r$ and $\mathbf{SAFE}^t$, we use the triangle inequality as

$$M = \frac{<\mathbf{SAFE}^r, \mathbf{SAFE}^t>}{<|\mathbf{SAFE}^r|, |\mathbf{SAFE}^t|>} \in \mathbb{C} \qquad (2)$$

with $M \in \mathbb{C}$. The argument $\angle M$ represents the angle between $\mathbf{SAFE}^r$ and $\mathbf{SAFE}^t$ (expected to be zero when the symmetry patterns detected coincide for reference and test feature vectors). The confidence is given by $|M|$, and it is expected to be $|M| \leq 1$ due to the triangle inequality [6]. To include confidence into the angle difference, we use $MS = |M| \cos \angle M$. The resulting score $MS \in [-1, 1]$ is equal to 1 for coinciding symmetry patterns in the reference and test vectors (full match).

*D. Rotation Invariance*

A beauty of the proposed method is that detection of intricate patterns is done directly in Cartesian coordinates via scalar product with filter $\psi_{nk}$. Transformations are not applied to the input image, but are implicitly encoded in the utilized filters. The descriptor also allows for rotation invariance without rotating the original image, nor the orientation map. Rotation can be efficiently achieved in the feature space instead by multiplication with a constant scalar as $SAFE'_{nk} = e^{2i(n+2)\varphi} SAFE_{nk}$ [11], which equates to rotate the original image with angle $\varphi$.

## III. BASELINE PERIOCULAR SYSTEMS

We test our system against the most widely used features in periocular research [2]: HOG [7], LBP [12], and SIFT [9]. In **HOG** and **LBP**, the periocular image is decomposed into non-overlapped square regions. Then, HOG and LBP features are extracted from each block. In HOG, the gradient orientation and magnitude are computed pixel-wise using the $[-1, 0, 1]$ and $[-1, 0, 1]^T$ kernels. The histogram of orientations is then built, with each bin accumulating corresponding gradient magnitudes. In LBP, each pixel is assigned a 8-bits label by thresholding the pixel intensity with each pixel in the 3×3 neighborhood. The binary labels of a block are then converted into decimal values and accumulated into an histogram. Both HOG and LBP are quantized into 8 different values to construct an 8 bins histogram per block. Histograms from each block are then normalized to account for local illumination and contrast variations, and finally concatenated to build a single descriptor of the whole periocular region. Matching with HOG and LBP can be done by simple distance measures.

Regarding the **SIFT** descriptor, key points extraction is done first using difference of Gaussians (DOG) functions in the scale space. A key point feature vector of dimension $4 \times 4 \times 8 = 128$ is then obtained by computing 8-bin gradient orientation histograms (relative to the dominant orientation to achieve rotation invariance) in 4×4 sub-regions around the key point. The recognition metric is the number of matched key points between two images. Since the number of key points is not necessarily the same in each image, the resulting score may be normalized by some function of the number of key points of the two images. To remove spurious matchings, we impose additional constraints to the angle and distance of matched key points [5].

## IV. DATABASE AND PROTOCOL

We employ the UBIPr periocular database [13]. UBIPr was acquired with a CANON EOS 5D digital camera in two sessions, with distance, illumination, pose and occlusion variability. The distance varies between 4-8m in steps of 1m, with resolution from 501×401 pixels (8m) to 1001×801 (4m). For our experiments, we select 1,718 frontal-view images from 86 individuals (corresponding to users with two sessions), thus having 86×2=172 available eyes. Two images are available per eye and per distance, resulting in 172×2=344 images per distance (except in D5, which has data from one user missing, thus having 342 images). All images have been annotated manually, so the radius and center of the pupil and sclera circles are available, which are used as input for the experiments. Images of each distance group are resized via bicubic interpolation to have the same sclera radius (we choose as target radius the average sclera radius $R_s$ of each distance group, given by the ground truth). We use the sclera for normalization since sclera is not affected by dilation (as it is the case of the pupil). Then, images are aligned by extracting a square region of $7.6R_s \times 7.6R_s$ around the sclera center. This size is set empirically to ensure that all available images have sufficient margin to the four sides of the sclera center. A summary of the available data is given in Table IV, and an example of normalized images is shown in Figure 3.

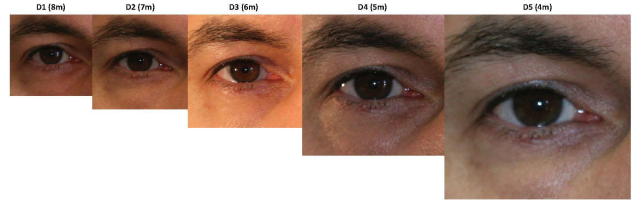

Fig. 3. Example of images from the UBIPr database captured at different distances (relative scale difference between images also shown).

|  | D1 | D2 | D3 | D4 | D5 |
|---|---|---|---|---|---|
| distance | 8m | 7m | 6m | 5m | 4m |
| images | 344 | 344 | 344 | 344 | 342 |
| average $R_s$ | 39.16 | 45.61 | 54.38 | 67.84 | 88.95 |
| image size | 299×299 | 347×347 | 415×415 | 517×517 | 677×677 |

TABLE I
DATABASE SPECIFICATION.

We carry out verification experiments, with each eye considered a different user. Enrolment images captured at distance $Di$ are matched against test images captured at distance $Dj$ ($i, j \in \{1, 2, 3, 4, 5\}$). For genuine matchings, we match all images of a user among them, avoiding symmetric matches. This results in 4,290 user scores. Concerning impostor experiments, the first image of a user is used as enrolment image, and it is matched with the second image of the remaining users, resulting a total of 439,480 scores.

Some fusion experiments are also done between different matchers. We use linear logistic regression fusion. Given $N$ matchers which output the scores ($s_{1j}, s_{2j}, ... s_{Nj}$) for an input trial $j$, a linear fusion of these scores is: $f_j = a_0 + a_1 \cdot s_{1j} + a_2 \cdot s_{2j} + ... + a_N \cdot s_{Nj}$. The weights $a_0, a_1, ... a_N$ are trained via logistic regression [3]. We use this trained fusion approach because it has shown better performance than simple fusion rules (like the mean or the sum rule) in previous works [3]. Nonetheless, this is a weighted sum rule, though the coefficients are optimized by a specific rule.

## V. RESULTS

The SAFE descriptor extracts orientation fields at different scales (Section II-A). Here we use $\sigma = \{1, 2, 3, 4\}$ with images at distance D1 (8m), scaling the value of $\sigma$ accordingly at other distances in proportion to the increase of the average $R_s$. The example of Figure 2 (left) corresponds to an image of distance D1 and the four values of $\sigma$ indicated above. We also employ $N_h = 9$ different families of symmetries (from $m = -4$ to 4), and $N_f = \{2, 3, 4, 5\}$ annular regions. The four SAFE features extracted at different $\sigma$ are simply grouped into a unique complex feature vector, which is used as identity model. For consistency, HOG, LBP and SIFT features are extracted from the same ROI as the SAFE system. The grid employed with LBP and HOG has 8×8=64 blocks, although the four corner blocks fall completely outside of the ROI, therefore features are extracted only from 60 blocks. Table II indicates the size of the feature vector for a given periocular image with the different matchers employed.

| system | | size | data type |
|---|---|---|---|
| SAFE | $N_f = 2$ | 4×2×9=72 | complex number |
| | $N_f = 3$ | 4×3×9=108 | complex number |
| | $N_f = 4$ | 4×4×9=144 | complex number |
| | $N_f = 5$ | 4×5×9=180 | complex number |
| LBP, HOG | | 60×8=480 | real number |
| SIFT | | 128 per key point | real number |

TABLE II
SIZE OF THE FEATURE VECTOR FOR EACH MATCHER.

### A. Individual systems

The performance of the SAFE periocular system is given in Figure 4. We test the system both with and without rotation compensation. Compensation is done by 'rotating' the test image in steps of 1 degree within the range $\varphi = \pm 15°$ (see Section II-D), and selecting the maximum $MS$ score as the best match between the two templates. It can be observed that the choice of $N_f$ has not a significant impact in the performance, at least in the range of values tested here. The only exception is $N_f = 2$, which shows a slight decrease in FRR at low FAR. The best cases both with and without rotation compensation can be marginally attributed to $N_f = 4$. It is also observed (right plot) that rotation compensation does not have appreciable effects in the performance. We attribute this to the fact that the images employed are captured in upright frontal position.

The performance of the baseline matchers is given in Figure 5. We test both the Euclidean and the $\chi^2$ distance with LBP and HOG, with $\chi^2$ giving better performance. The Euclidean distance is the usual choice in previous periocular studies [14], whereas the $\chi^2$ distance has been observed to perform better with normalized histograms using other type of features, e.g. [1], due to giving more weight to regions with lower probability. Regarding the SIFT matcher, the number of matched key points between image pairs is normalized in our experiments either with the average number of detected key points in the two images, or with the minimum. Due to scale variations in the database employed, the number of detected key points is smaller as the acquisition distance increases. As a result, the minimum normalization is less sensitive when matching images captured at different scales, as our results in Figure 5 shows.

In the remainder of this paper, we employ the best configurations of each individual matcher, as given by the results of this section. EER values of these configurations are given in Table III, top. The SAFE matcher stays on top just behind HOG, whereas EER of the two other matchers is around a 3% higher or more. It is interesting to note that the SAFE feature vector has 144×2=288 real numbers (Table II), which is 60% the size of LBP or HOG vectors. SIFT vectors on the other hand are much higher, with 128 real numbers per key point, and dozens of key points per image. We could even reduce the size of SAFE vectors to 108×2=216 (45% the size of LBP or HOG) or 72×2=144 (30%) without a big sacrifice in performance.

### B. System combination

We then perform fusion experiments of the available matchers (Table III). We have tested all possible fusion combinations. From observations of Table III, the biggest performance improvement occurs after the fusion of two systems, with the inclusion of three of all systems producing only marginal improvements. The best EER is given by the fusion of all systems, although this should not be taken as a general rule, since the best fusion combination may not always involve all the available systems [4]. We also report the EER of the SIFT+LBP+HOG combination from a previous study using the same database with only frontal images [13]. It should be noted that another setup of the matchers is used in [13], including different ROI selection, a different fusion scheme and the use of Euclidean distance with LBP and HOG. A reason of the better performance obtained in our experiments could be the superior performance of the individual matchers due to the choice of $\chi^2$ distance, but this would need to be verified with additional experiments.

### C. Scale variation

We lastly analyze periocular recognition performance under image scale variations. Figure 6, left, shows the verification performance when both enrolment and test images are captured at the same distance (scale-wise), whereas the right plot shows the performance when enrolment and test images are captured at different distances (cross-scale). The figure includes results both of the individual matchers and of the best fusion combinations of Table III.

From Figure 6, left, we observe that each matcher performs better at a different scale. All matchers except SIFT perform the worst at the furthest acquisition distance (D1), with LBP showing a significant degradation. SAFE and LBP perform best at D4, whereas HOG performs best at D5, and these three matchers also show a steady improvement in performance as the acquisition distance is reduced. SIFT on the other hand shows a more unpredictable behaviour, with no correlation between distance and performance. The best fusion of two systems (SAFE+SIFT) seems to be influenced by this unpredictable behaviour of SIFT, although the fusion of three and four systems do shows a more clear correlation between distance and performance.

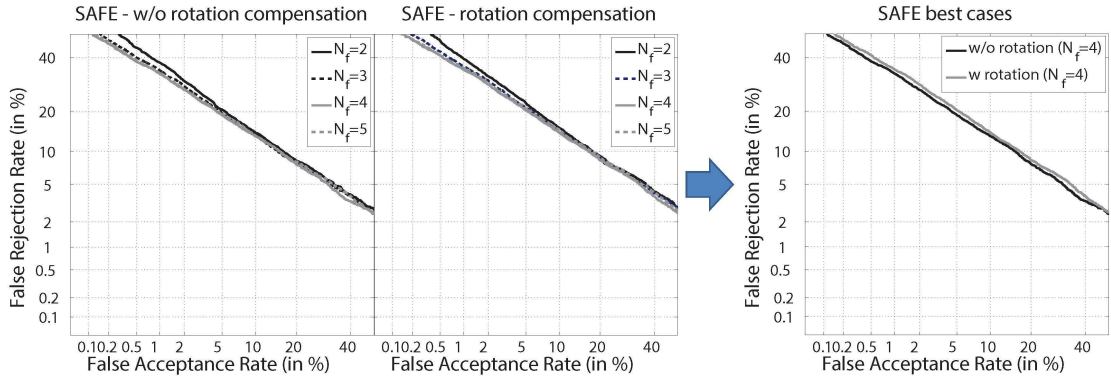

Fig. 4. Verification results of the SAFE matcher.

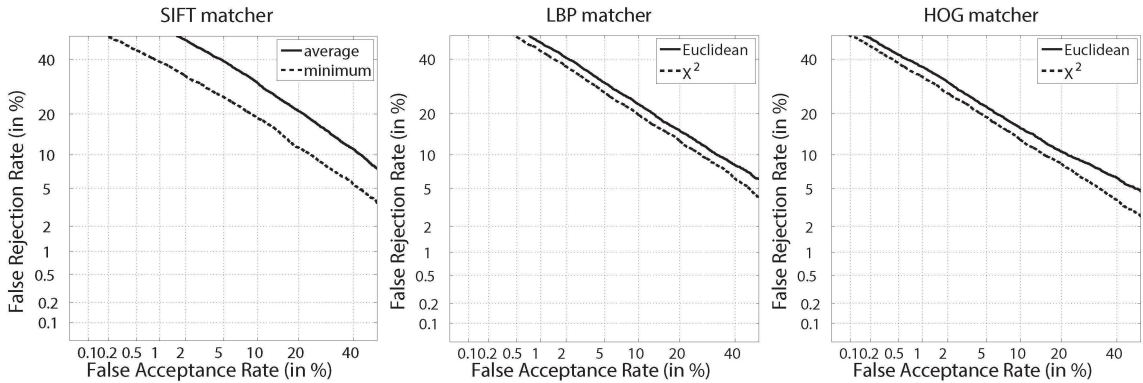

Fig. 5. Verification results of the SIFT, LBP and HOG matchers.

| systems | EER | variation |
|---|---|---|
| SAFE | 12% | - |
| SIFT | 14.9% | - |
| LBP | 15.5% | - |
| HOG | 11.8% | - |
| SAFE+SIFT | **8.9%** | (-25%) |
| SAFE+LBP | 10.9% | (-9%) |
| SAFE+HOG | 10% | (-15%) |
| SIFT+LBP | 9.6% | (-37%) |
| SIFT+HOG | **9%** | (-24%) |
| LBP+HOG | 11.5% | (-2%) |
| SAFE+SIFT+LBP | 8.3% | (-31%) |
| SAFE+SIFT+HOG | **8.2%** | (-30%) |
| SAFE+LBP+HOG | 9.8% | (-17%) |
| SIFT+LBP+HOG | 8.4% | (-29%) |
| all | **7.9%** | (-33%) |
| SIFT+LBP+HOG [13] | 16% | - |

TABLE III
VERIFICATION RESULTS (EER) FOR THE INDIVIDUAL MATCHERS AND FOR THEIR DIFFERENT FUSION COMBINATIONS. THE RELATIVE EER VARIATION WITH RESPECT TO THE BEST INDIVIDUAL SYSTEM IS GIVEN IN BRACKETS. A PREVIOUS STUDY MAKING USE OF THE SAME DATABASE IS ALSO REPORTED.

Regarding cross-scale experiments (Figure 6, right), it is interesting to note that all matchers perform the worst when enrolment and test images are captured at the same distance. A plausible explanation is score misalignment, since this case encompasses matching of images captured at 8 meters among themselves, plus matching of images captured at 7 meters among themselves, etc. The SIFT matcher is the most affected by this effect, since its matching score is a function of the number of detected key points, whose amount is not the same across different distances. The fusion of two systems (SAFE+SIFT) shows a significant improvement w.r.t any individual system, further improved by the fusion of three systems. It is also worth noting the somewhat stable performance between distance variations of 1 and 5 (except the peak at 4). This means that the matchers are able to cope relatively well with differences in acquisition distance between enrolment and test images.

## VI. CONCLUSIONS

This paper describes a new periocular system based on detection of local symmetry patterns (Figure 1) in annular rings around image key points. We employ as unique key point the sclera center, highlighting the unique object-like properties that concentrate to this anchor point of the eye. Using one key point only also allows to keep the size of the feature vector small, while using the sclera as reference mitigates the effect of pupil dilation. The sclera is also easier to detect in visible range, which is the illumination that best suits the periocular modality, both in terms of performance and of the difficulty (or even impossibility) of using near-infrared cameras in unconstrained environments [2].

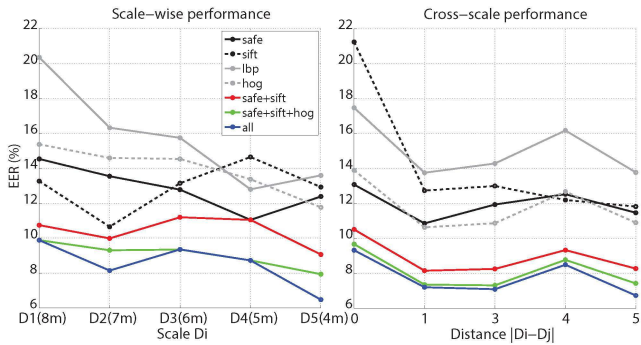

Fig. 6. Verification results (EER) for the scale variation experiments.

The system is evaluated with a database of frontal periocular images captured with a digital camera. We test a different number of annular rings, with no significant impact on the performance (at least in the range of values used). This allows to achieve even further reductions in the size of the feature vector. We also observe that rotation compensation does not have any effect on the performance, which could be due to the use of images captured in upright position. The system proposed here is compared with reference periocular descriptors [14], encompassing Local Binary Patterns (LBP), Histogram of Oriented Gradients (HOG), and Scale-Invariant Feature Transform (SIFT) key points. Our system is able to obtain top performance with a much smaller feature vector, with further improvements obtained via fusion of the available systems. We also evaluate the impact of the acquisition distance in the performance, with the majority of matchers showing better performance as the acquisition distance is reduced. This effect is not observed in previous studies with the same database [13], although authors here included non-frontal images in the experiments. In addition, the matchers are able to cope well when enrolment and test images are not captured at the same distance.

Future work includes testing the proposed method with difficult images with other illumination, for example near-infrared portal data [8] or night-vision [17]. Another area where we believe that we can contribute with our system is heterogeneous scenarios involving cross-sensor [16] or cross-spectral [17] recognition, which will appear for example if people is allowed to use their own sensors (e.g. smartphones) or images from different spectra have to be imperatively matched (e.g. forensics). Testing the limits of our system to increased gaze angle or off-centered images is also being considered.

ACKNOWLEDGMENT

F. A.-F. thanks the Swedish Research Council for funding this research. Authors also acknowledge the CAISR program of the Swedish Knowledge Foundation.